\documentclass[conference]{IEEEtran}
\usepackage{multirow}

\IEEEoverridecommandlockouts
\usepackage{graphicx}
\usepackage{amsmath}
\usepackage{amssymb}
\usepackage{multirow}
\usepackage{booktabs}
\usepackage{url}
\usepackage[table]{xcolor}
\usepackage{xcolor}
\usepackage{pgfplotstable}
\usepackage{siunitx}

\def\BibTeX{{\rm B\kern-.05em{\sc i\kern-.025em b}\kern-.08em
		T\kern-.1667em\lower.7ex\hbox{E}\kern-.125emX}}
\def\cca#1{%
	\pgfmathsetmacro\calc{100-(#1-0)*100/(1-0)}%
	\edef\clrmacro{\noexpand\cellcolor{green!\calc}}%
	\clrmacro%
	\ifdim \calc pt>50pt\color{black}\fi{#1}%
}

\def\ccag#1{%
	\pgfmathsetmacro\calc{25}%
	\edef\clrmacro{\noexpand\cellcolor{green!\calc}}%
	\clrmacro%
	\ifdim \calc pt>50pt\color{black}\fi{#1}%
}

\def\ccar#1{%
	\pgfmathsetmacro\calc{25}%
	\edef\clrmacro{\noexpand\cellcolor{red!\calc}}%
	\clrmacro%
	\ifdim \calc pt>50pt\color{black}\fi{#1}%
}
\begin{document}

	\title{Hand Gesture Classification on Praxis Dataset: Trading Accuracy for Expense}
	
		\author{\IEEEauthorblockN{ 
				Rahat Islam, Kenneth Lai and Svetlana N. Yanushkevich
			}
			\IEEEauthorblockA{
				\textit{Biometric Technologies Laboratory, Department of Electrical and Software Engineering,} \textit{University of Calgary, Canada} \\
				Web: http://www.ucalgary.ca/btlab, E-mail: \{rahat.islam, kelai, syanshk\}@ucalgary.ca}\newline
		}

	\maketitle
	
	\IEEEpubid{
		\begin{minipage}{\textwidth}\ \\[70pt]
					\textbf{\footnotesize{{\fontfamily{ptm}\selectfont Digital Object Identifier 10.1109/IJCNN55064.2022.9892631\\978-1-7281-8671-9/22/\$31.00 \copyright 2022 IEEE}}}
		\end{minipage}
	}

	\begin{abstract}
		In this paper, we investigate hand gesture classifiers that rely upon the abstracted 'skeletal' data recorded using the RGB-Depth sensor. We focus on 'skeletal' data represented by the body joint coordinates, from the Praxis dataset. The PRAXIS dataset contains recordings of patients with cortical pathologies such as Alzheimer's disease, performing a Praxis test under the direction of a clinician.  In this paper, we propose hand gesture classifiers that are more effective with the PRAXIS dataset than previously proposed models. Body joint data offers a compressed form of data that can be analyzed specifically for hand gesture recognition. Using a combination of windowing techniques with deep learning architecture such as a Recurrent Neural Network (RNN), we achieved an overall accuracy of 70.8\% using only body joint data. In addition, we investigated a long-short-term-memory (LSTM)  to extract and analyze the movement of the joints through time to recognize the hand gestures being performed and achieved a gesture recognition rate of 74.3\% and 67.3\% for static and dynamic gestures, respectively. The proposed approach contributed to the task of developing an automated, accurate, and inexpensive approach to diagnosing cortical pathologies for multiple healthcare applications. 
	\end{abstract}
	
	
	%
	\section{Introduction}\label{sec:introduction}
	
	Computer vision  has made a huge contribution to the field of hand gesture recognition from video. To humans, the use of hand gestures is an indispensable way for us to communicate with others. With increased interactions between computers and humans, researchers are delving into the  ways in which hand gestures could also be used as a way for humans to interact with computers. For instance, paper \cite{Gupta2016Towards} looked into how hand gestures can be used as a way for drivers to signal to their vehicle when certain services are needed. By using hand gestures to communicate with the vehicle, the driver can keep his eyes on the road and handle the vehicle safely in a more convenient way. The use of hand gesture recognition with human-computer interfaces is also being investigated for the hearing impaired, people who rely heavily on hand gestures to communicate \cite{soni2016online}. Gesture recognition has also found its way into the medical diagnostics field, and may potentially have a large impact on the creation of autonomous systems which are capable of identifying different cortical pathologies, simply by visually 'reading' a patient's hand gestures in a controlled setting. The last topic is especially important for this paper, as our goal is to investigate and create machine learning models which are effective in reading hand gestures of patients with  Alzheimer’s disease (AD). AD is a condition that affects motor movements of the body. Studies show that loss in motor movements correlates with the severity of AD \cite{Chandra2015apraxias}. AD can also be  diagnosed by using a so-called Praxis test. Praxis is defined as the ability to perform skilled movements in a non-paralytic limb. Apraxia is the term used for the inability to perform a skilled movement with a limb despite having no sensory or motor deficiencies. The American Psychiatric Association finds that the Praxis test is an indicative test for cortical pathologies like Alzheimer’s \cite{American2000dsm}. Despite that, most clinicians and examiners refrain from using the exam on their patients. This is due to the shortage of specialists who can conduct and evaluate the test. Even among the specialists, the instructions are not standardized, thus the test is not objective enough. Therefore, there is a need to systematically perform the test and have an automated system that can accurately evaluate apraxia for any cortical pathologies.
	
	In recent years, the development of depth sensors like the Kinect sensor \cite{WangBeyond2013} has enabled people to observe hand movements with high accuracy without the need for wearable sensors which can be hard to use for inexperienced users. Such sensors can pick up depth, RGB, and body joint information all at once. Body joints in particular have grown in popularity in their effectiveness to assist in computer vision tactics for gesture recognition \cite{ionescu2005dynamic}.  A Convolution Neural Network + Recurrent Neural Network approach is proposed in \cite{lai2018cnn+} to extract spatial-temporal information to recognize dynamic hand gestures using depth and body joint information.  Newer learning techniques involving an ensemble of different models \cite{lai2020ensemble}, Temporal Convolution Networks 
	\cite{lea2017temporal, lai2021capturing}, multi-radii circular signature \cite{sahana2022mrcs}, 3-D CNNs \cite{sharma2021asl}, feature fusion RCNN \cite{xu2022novel}, and elastic semantic network \cite{li2021else} have been proposed for gesture/action recognition.
	
	To tackle the challenge of designing an autonomous system used in for medical diagnostics of AD, the PRAXIS dataset was created by Negin et. al. \cite{Negin2018praxis}. To record the data, patients were each asked to perform a set number of various hand gestures. Some of the patients were healthy patients, while others had pathologies such as in the case of AD, which inhibited certain movements in their limbs. Besides the recorded gestures for each patient, the dataset also contains information on whether the movement was done successfully or not. The creators of the dataset used this information to not only classify various gestures but also to evaluate whether a person's inability to perform different movements is indicative of a cortical pathology. The accuracy of gesture classification proposed in \cite{Negin2018praxis} using only skeletal data achieved  63.5\%.
	
	The goal of our paper is to create advanced machine learning models of a gesture classifier of the gestures from the PRAXIS dataset, that would be accurate but computationally inexpensive by using only 'skeletal' data. Note that the Praxis dataset applies for assessing  medical-related motor function abnormalities, and was not created to serve as a dataset for regular gesture recognition. Therefore, this dataset has not been researched as much as other datasets. We set on a mission to prove that advanced machine learning models such as deep neural networks can provide greater accuracy on the PRAXIS dataset. Our proposal revolves around the use of Long-Short Term Memory (LSTM) and Temporal Convolutional Network (TCN). In addition, we propose to apply certain pre-processing techniques to improve gesture recognition accuracy. Note that while the gesture classifier alone would not suffice for a fully functional automated system that can perform the Praxis test, this classifier is the foundation for developing such a system.
	
	The structure of the paper is as follows. We first describe the Praxis dataset in more depth, and explain existing gesture classification models for this dataset. Next, we introduce our classification models and the methodology behind them. Lastly, we report the experimentation results and future works.
	
	\section{Dataset and Related Work}\label{sec:framework}

	Since the development of inexpensive RGB-depth sensors such as Kinect v2, many advanced automated technologies based on sensory data were developed. These technologies include the detection and identification of human behavioral biometrics, including human body action and gesture recognition. The development of high accuracy RGB-depth sensors   has enabled people to observe hand movements with high accuracy without the need for contact-based sensors. Such sensors can pick up depth, RGB, and body-joint structures, also called 'skeletal' images, all at once. 'Skeletal' images in particular have grown in popularity in their effectiveness to assist in computer vision tactics for body pose estimation \cite{WangBeyond2013} and hand gesture recognition \cite{ionescu2005dynamic}.
	This also benefits healthcare applications. In particular, using Kinect v2, the Praxis dataset \cite{Negin2018praxis} was created for the purpose of recording patients doing the Praxis test known to detect cortical pathologies via analysis of upper body movements. The data was also used to create hand gesture classification models to evaluate the Praxis test objectively.  
	
	\subsection{PRAXIS Dataset}
	
\begin{figure}[!ht]
		\begin{center}
			\includegraphics[width=0.24\textwidth]{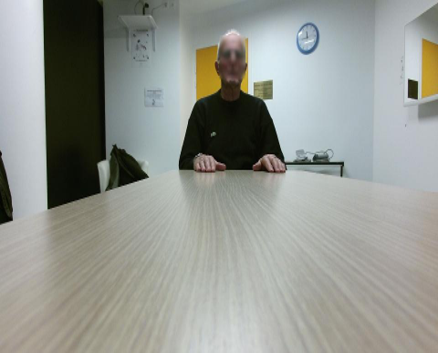}
			\includegraphics[width=0.24\textwidth]{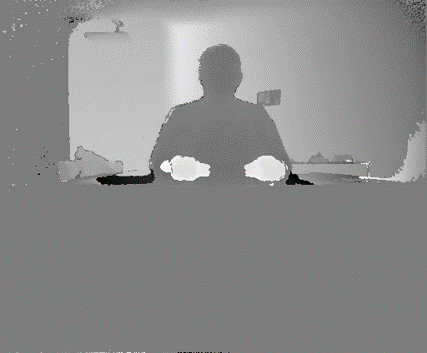}
			\includegraphics[width=0.24\textwidth]{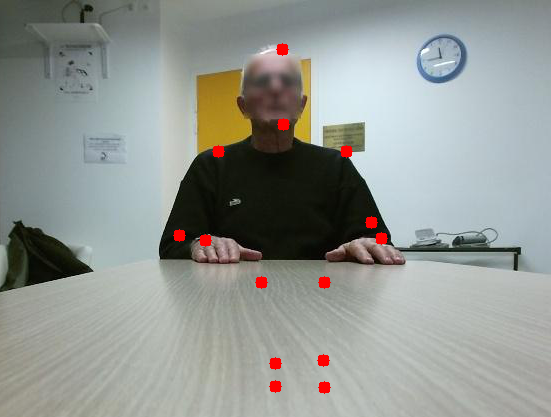}
			\caption{Depth image (top left), RGB image (top right) and body joint data points (bottom) are all used to record the motion of patients.}
			\label{fig:Biases}
		\end{center}
	\end{figure}

	In this section, we describe the Praxis dataset \cite{Negin2018praxis} in-depth and look at other gesture classification models that utilized this dataset. Table \ref{tab:gestures1} illustrates the 29 different gestures performed by each patient in the dataset. Before each gesture is performed, a clinician performs the movement, and the patient is asked to imitate it. The way the video is captured is such that only the upper body of the patient can be seen. Therefore, all the gestures are essentially hand gestures. In the dataset, 56 patients perform the shown gestures. An important distinction to make among the different gestures is that some are static gestures, while others are dynamic. Static hand gestures can be considered simply as still poses where the gesture itself is in a fixed position. On the other hand, dynamic gestures include motion in the gesture itself, often making the movement more complex and harder to identify. In this dataset, patients are each asked to perform a set number of different hand gestures. Some of the patients are healthy patients, while others have pathologies like AD which inhibit certain movements in their limbs. Not only does the dataset record the gestures for each patient, but it also shows whether the movement was done successfully or not. The creators of the dataset used all this information to not only classify different gestures but also to evaluate whether a person's inability to perform different movements is indicative of a cortical pathology. Fig. \ref{fig:Biases} sample images (RGB and depth) from the Praxis dataset.
	
	\begin{figure*}[!ht]
		\begin{center}
			\begin{tabular}{ccccc}
				\includegraphics[width=0.15\textwidth]{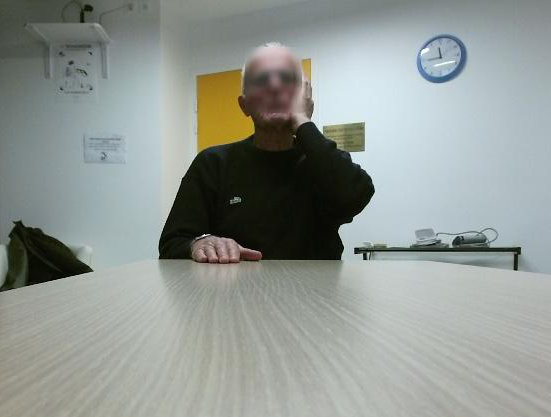}
				&
				\includegraphics[width=0.15\textwidth]{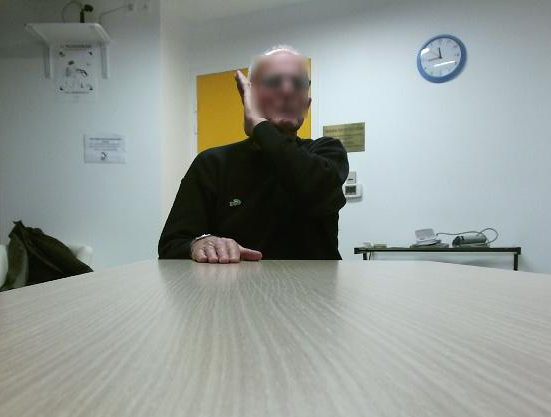}
				&
				\includegraphics[width=0.15\textwidth]{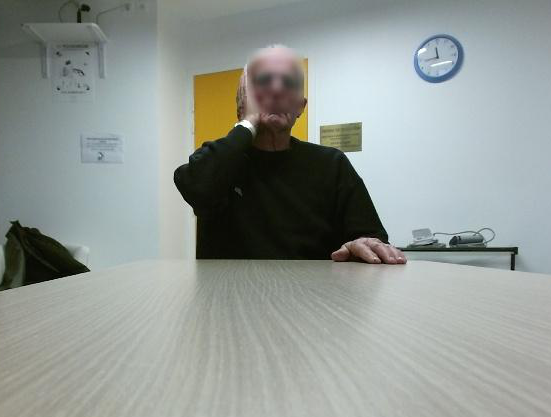}
				&
				\includegraphics[width=0.15\textwidth]{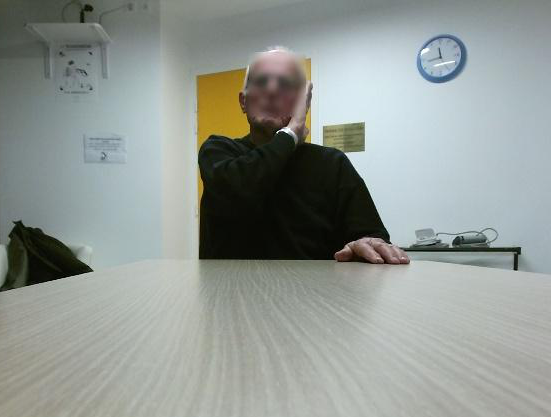}
				&
				\includegraphics[width=0.15\textwidth]{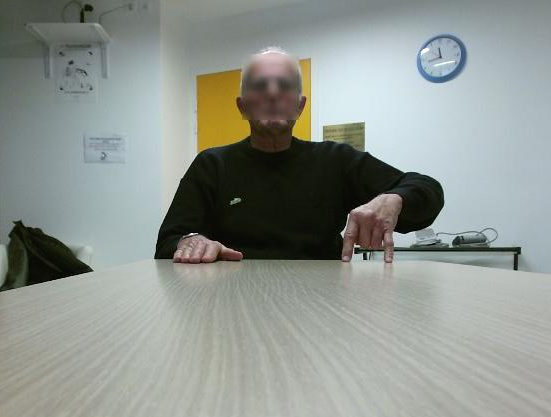}
				
				\\
				A1\_1& A1\_&A1\_3&A1\_4&A1\_5 \\
			\end{tabular}
			\caption{A sample of what different poses look like in the Praxis dataset. A1\_1: Left hand on left ear, A1\_2: Left hand on right ear, A1\_3: Right hand on right ear, A1\_4: Right hand on left ear, and A1\_5: Index and baby finger on table.}
		\end{center}
	\end{figure*}
	\begin{table}
		\begin{center}
			\caption{All Hand Gestures With Descriptions}\label{tab:gestures1}
			\begin{tabular}{ccc}
				\hline
				ID& Type & Description\\
				\hline
				A1\_1 & Static & Left hand on left ear \\
				A1\_2 & Static & Left hand on right ear \\
				A1\_3 & Static & Right hand on right ear  \\
				A1\_4 & Static & Right hand on left ear \\
				A1\_5 & Static & Index and baby finger on table \\
				A2\_1 & Static & Stick together index and baby fingers \\
				A2\_2 & Dynamic & Hands on table, twist toward body \\
				A2\_3 & Static &Bird\\
				A2\_4 & Static & Diamond \\
				A2\_5 & Static & ring together \\
				S1\_1 & Static & Do a military salute \\
				S1\_2 & Static & Ask for silence \\
				S1\_3 & Static & Show something smells bad \\
				S1\_4 & Dynamic & Tell someone is crazy \\
				S1\_5 & Dynamic & Blow a kiss \\
				S2\_1 & Dynamic & Twiddle your thumbs \\
				S2\_2 & Static & Indicate there is unbearable noise \\
				S2\_3 & Static & Indicate you want to sleep \\
				S2\_4 & Static & Pray \\
				P1\_1 & Dynamic & Comb hair \\
				P1\_2 & Dynamic & Drink a glass of water \\
				P1\_3 & Dynamic & Answer the phone \\
				P1\_4 & Dynamic & Pick up a needle \\
				P1\_5 & Dynamic & Smoke a cigarette \\
				P2\_1 & Dynamic & Unscrew a stopper \\
				P2\_2 & Dynamic & Play piano \\
				P2\_3 & Dynamic & Hammer a nail \\
				P2\_4 & Dynamic & Tear up a paper \\
				P2\_5 & Dynamic & Strike a match \\
				\hline
			\end{tabular}
		\end{center}
	\end{table}
	
	Each frame that captures different movements in the dataset comes in three different forms: an RGB image, a depth image, and a set of data points containing the coordinates of 14 body joints (this is also called a 'skeletal' data). All three types of data can be used for gesture classification as shown in \cite{Negin2018praxis}. 
	
	\subsection{Related Work}
	
	In this section, we describe the approaches to  gesture classification using the RGB-Depth data.
	
	In the paper by the creators of the Praxis dataset, \cite{Negin2018praxis}, the four classifiers were investigated: the skeletal-based method, multi-modal fusion, local descriptor-based method, and the deep learning-based approach.
	
	\textbf{Skeletal-based method:} Using only skeletal data, joint angle and Euclidean distance features are formed to create the global appearance of the poses \cite{Negin2018praxis}. Furthermore, a temporal window is used to catch the temporal dependencies of each action. The data is then passed into an SVM for classification which reaches only 63.5\% accuracy.
	
	\textbf{Multi-modal fusion:} In addition to using the skeletal-based method, this method also extracts deep VGG features and a VGG deep neural network from RGB images to get more information on the shape of the hand \cite{Negin2018praxis}. Before classification, a late fusion scheme is used to form class probabilities. This method is used in \cite{Negin2018praxis} and the classifier  reaches 67.6 \% accuracy using only RGB(VGG) and skeletal data.
	
	\textbf{Local descriptor based method:} The method incorporates action recognition techniques with improved dense trajectories similar to the approach proposed in \cite{WangAction2013}. The feature extraction step is based on a Fisher vector encoding scheme. 4.	This approach extracts action descriptors from images and uses a linear SVM classifier which reaches a 72.9\% accuracy \cite{Negin2018praxis}.  
	
	\textbf{Deep learning-based method:} A Convolutional Neural Network (CNN) is coupled with an LSTM to form this hand gesture classifier. While a CNN is very strong at learning the dynamics of the hand movements, an LSTM network serves the purpose of learning the temporal dependencies of each action. This allows the CNN and LSTM network to work hand in hand to assist in a strong classifier. The deep learning classifier uses skeletal data as well as RGB images, but not the depth data \cite{Negin2018praxis}. This deep learning method has the highest accuracy of 84.7\%, but at the expense of using RGB images which is computationally expensive and possibly harder to use in real life scenarios. RGB images can cause privacy issues as it reveals a persons identity. On the other hand, skeletal information makes it harder to identify a person. Furthermore, classification of RGB images also suffer in darker settings while the sensors which collect joint information wouldn't suffer to the same degree.
	
	The experimental results shown in \cite{Negin2018praxis} indicates that a deep learning-based method provides superior results compared to the other methods and that body joint information has immense potential. Inspired by these results, we choose to use an RNN, a deep learning architecture, for gesture recognition and choose to focus on using only body joint information.  Descriptor techniques and fusion methods involving multiple data types serve as a comparison to our proposed approach.

	\section{Proposed Approach}\label{sec:framework}
	In this section, we consider the framework behind the design of our machine learning models. We decided to split our data into three folds. Fold one contains the data for patients 1-15, Fold two holds the data for patients 16-35, and Fold three combines the data for patients 36-55. Since each patient performed each of the 29 different actions, the 3-fold split had a similar amount of representation of each action in each fold. Thus, we did not have to worry about one fold being less balanced than another. However, some patients had unsuccessful attempts at certain gestures. Since we want to classify gestures, an incorrect gesture labeled as a certain action would mislead our machine learning model. Therefore, we removed any incorrect gestures from the data.
	
	During the initial stages of development of our classification model, we decided to use only skeletal joints due to its effective purpose of pin-pointing specific joints on the body as well as its being a much smaller size than the depth and RGB images. This enables the machine learning model to not be very computationally expensive with the large dataset, in comparison to using the other images for data. A similar approach was used in one of the methods by \cite{Negin2018praxis}. In \cite{Negin2018praxis}, the highest accuracy achieved with skeletal joints alone was 63.5\%, making it an important baseline for our experiments.  Next, we will talk about different methods that were tested and compared in the process of designing the hand gesture classifier.
	
	\subsection{Data management within each frame}
	
	The skeletal data for each frame is represented by a $5 \times 14$ matrix of numbers. The 14 columns were formed by the 14 different joints detected using Deepcut, a CNN-based body part detector \cite{pishchulin2016deepcut}. The first two rows were the $x$ and $y$ coordinates of each joint. The third row was confidence values for each joint, whereas the last two rows provided additional data for the users. For our experiment, we  removed it from the input data, after having two trials: one with the last two rows were kept alongside the $x$ and $y$ coordinates, and another trial in which the last two rows were deleted. Since excluding the last two rows of skeletal data for each frame had no impact on the accuracy of our classifier, we proceeded with the removal of those rows.

	\subsection{Time windows for sequential data}
	Because a gesture is an action performed in time, the data we are working with is sequential. In other words, one frame of a gesture is meaningless, but a collection of frames can be used to capture the action. Thus, for our classifier, the input would be a collection of $x$ number of frames, and a sliding window approach would be used to read in an entire action. Naturally, to capture all the frames of a full movement, each window would be shifted by one frame until the action was fully captured. The next step was to figure out how big each time window should be (how many frames), in order to yield  the best classification results. During the experiments, we started at a time window size of two frames and incremented that number by a power of two,  until we had 512 frames per action. Note, that some actions lasted only about 50 frames. To make sure such action would still fit in a window larger than 50 frames, blank frames (zeros as data entry) were inserted at the beginning of the window. 
	
	\subsection{Data normalization}
	
	Before doing any normalization, we applied the Savgol filter to the data, to smoothen out the joints as suggested in \cite{gallagher2020savitzky}. Suppose ${x_{j}, y_{j}}, j= 1,..n,$ where $x_{j}$ are the x coordinates of each joint and $y_{j}$ are the corresponding y coordinates. The smoothened out y coordinates are treated with $m$ convolution coefficients, $C_{i}$. The coefficient values could be found in the table of selected convolution coefficients \cite{gallagher2020savitzky}. Thus the formula used in the Savgol filter for our data is as follows.  
	
	\[Y_{j} =\sum_{\frac{1-m}{2}}^{\frac{m-1}{2}}C_{i}y_{j+i},  \frac{m+1}{2}\leq j \leq\frac{m-1}{2}\]
	
	In our experiments, we chose to set $m = 5$ and smoothened the data as a quadratic polynomial. These details were needed in choosing the appropriate coefficient values.
	
	We have also resolved a challenge with data rotation due to each patient's location with respect to the camera, which means that the coordinates of the  patient's joints in the skeletal data have a reference point specific to the camera location. In other words, if a patient is shifted to the right or to the left from the camera, the $x$ values will all be skewed in comparison to another patient. For that reason, with each time window, we remapped the coordinates of the joints with respect to the chin node (coordinate point indicating the chin location) in the first frame of each time window. The coordinates were also calculated in four ways, to find which method would yield the best accuracies. Below, we will go over the first method in more detail, as well as the other four methods.
	
	Let $ x_{chin} $ and $ y_{chin} $ to be the $ x $ and $ y $ coordinates of the chin node in the first frame in a time window. Let $x_{new}$ and $y_{new}$ be the new coordinates for any joint in any frame within a time window, and $x_{original}$ and $y_{original}$ be the original $x$ and $y$ coordinates (as originally found in the dataset) of an arbitrary joint in any frame within the time window.
	
	\textbf{Method 1:} This method  alters the skeletal data by making all the coordinates in reference to the chin node of the first frame. The benefit of this is that the location of the patient with respect to the center of the camera does not affect the coordinates of each joint in the skeletal data:
	\[x_{new} = x_{original} - x_{chin},
	y_{new} = y_{original} - y_{chin}\]
	
	\textbf{Method 2:} Method two is similar to the first one, except that it also normalizes the coordinates with respect to the original coordinates of the chin node in the first frame. The equations are as follows:
	\[x_{new} = \dfrac{x_{original} - x_{chin}}{x_{chin}}, 
	y_{new} = \dfrac{y_{original} - y_{chin}}{y_{chin}}\]
	
\begin{figure}[!ht]
		\begin{center}
			\includegraphics[width=0.33\textwidth]{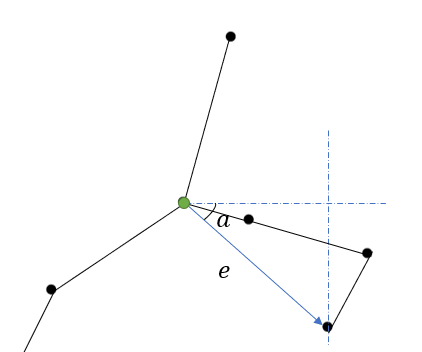}
			\caption{Demonstration of how polar coordinates are extracted from the skeletal joints; here,  the green dot represents the chin node of the first frame within the chosen time window.}
			\label{fig:polar}
		\end{center}
	\end{figure}

	\textbf{Method 3:} Method three differs from the other two methods; instead of using rectangular coordinates, it uses polar coordinates. Suppose $e$ is the Euclidean distance and $a$ be the angle. Note that the reference node is still the chin node from the first frame. Figure \ref{fig:polar} shows a visual representation of  how the polar coordinates are extracted from a skeletal data,
	\[e = \sqrt{(x_{original}-x_{chin})^{2} + (y_{original}-y_{chin})^{2}}\]
	\[a = arctan(\dfrac{y_{original}-y_{chin}}{x_{original}-x_{chin}})\]
	
	\textbf{Method 4 and Method 5:} Method 4 is a combination of methods 1 and  3, where both types of data points are passed in. Method 5  is a combination of methods 2 and  3. Note that the equations for the last two methods will not be shown to avoid redundancy.
	
	Each of these methods is used for experimentation in order to determine which method provides the best performance.  Depending on the method, the result will indicate which combination of pre-processing techniques is the best for gesture recognition.
	
	\subsection{Deep Learning Networks}
	
	Deep learning approach deploys a neural network with multiple layers to extract high-level features from raw data \cite{lecun2015deep}. In our experiments, we investigate two different types of neural networks: a Long Short Term Memory (LSTM) network \cite{hochreiter1997long} and a Temporal Convolutional Neural Networks (TCN) \cite{chen2020probabilistic}.	
	
	\subsubsection{LSTM}
	
	LSTM belongs to Recurrent Neural Networks (RNNs) \cite{hochreiter1997long}, a special class of neural networks. Unlike a standard feed-forward network, the RNN has a feedback mechanism that allows it to look at data as a sequence of points rather than a single point. Being able to handle sequential data is important for our experiments, since in order to understand a gesture, you have to watch it from beginning to end. Thus, the nature of the data itself is sequential. However, one problem with RNNs is that it struggles to handle long sequences due to their short-term memory. In other words, by the time an RNN finishes encoding the sequence, the initial data points will probably be lost. LSTMs use a gated mechanism to solve the short-term memory problem and thus are more effective in handling long sequential data, which is ideal for our experiments.
	
	\subsubsection{Temporal Convolutional Networks}
	
	CNNs have often been used with image classification, while LSTMs were used in the classification of sequential data. However, with recent advancements in the architecture of CNNs \cite{wang2016cnn},  CNNs have also started to make their appearance to process sequential data \cite{chen2020probabilistic}. The TCN is a class of 1-dimensional CNNs. Through the use of dilated causal convolutions as well as stacking such convolutional layers on top of each other, TCNs can handle long sequential data without being too computationally expensive. Furthermore, the gradient flow direction of a TCN differs from an LSTM such that this model will not suffer the vanishing gradient problem that RNNs and  LSTMs still face given long series of sequential data.
	
	\subsection{Number of Classifiers}
	
	In this section, we describe two models that operate differently based on the number of classes used: the multi-class classifier and the multi-class binary classifier. In \cite{Negin2018praxis},  a binary classification system was used to identify whether an image belongs to a certain class or not. This procedure had to be performed 29 times as there were 29 classes, and an average of the accuracies was calculated to get a total accuracy. Since we aim to compare the performance of our hand gesture classification model to the above, we follow the same evaluation process. Thus, we did a binary classification for each type of action/class and took the average of those to get an average accuracy.
	
	\begin{figure*}[!ht]
		\begin{center}
			\includegraphics[width=0.75\textwidth]{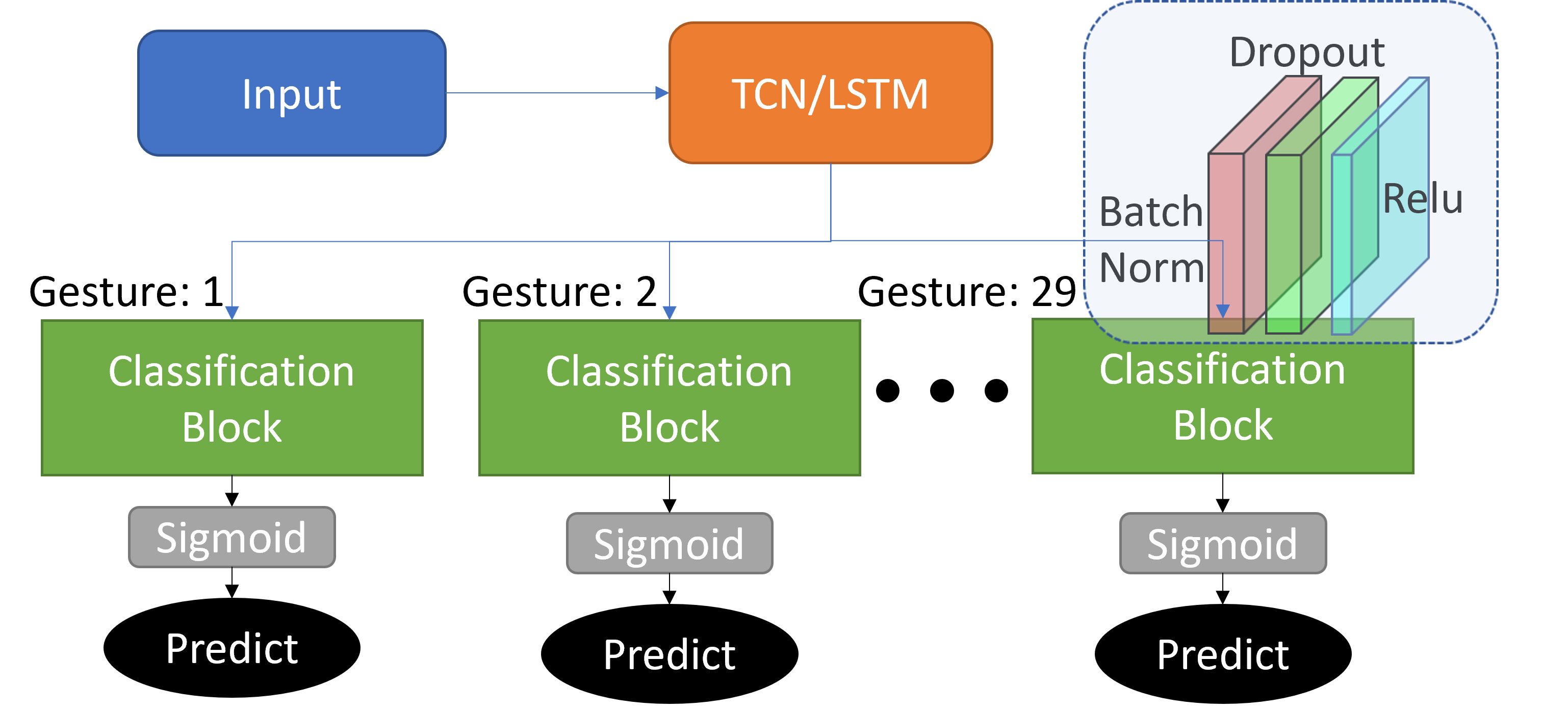}
		\end{center}
		\caption{A general architecture for our multi-class binary classification model.}
		\label{fig:multi}
	\end{figure*}
	
	The second method we consider is a traditional multi-class classification in which the classifier would read an image, and classify it to belong to one of the classes. Though this method was not used by \cite{Negin2018praxis}, we wanted to employ this method since the first one does not necessarily classify which class an action belongs to. For instance, when evaluating an action as A2-3, the binary classifier might find it positive, but the same action can be classified as A2-4. This does not provide a unique answer to which class the image belongs to. Using the multi-class classification, we would not have this problem as only one action can be classified into one class. It is worth mentioning that for the multi-class classifier, we had  separated  the classification for static hand gestures and dynamic hand gestures. Our average accuracy would be the average of the two types of gestures.
	
	\begin{figure*}[!ht]
		\begin{center}
			\includegraphics[width=0.75\textwidth]{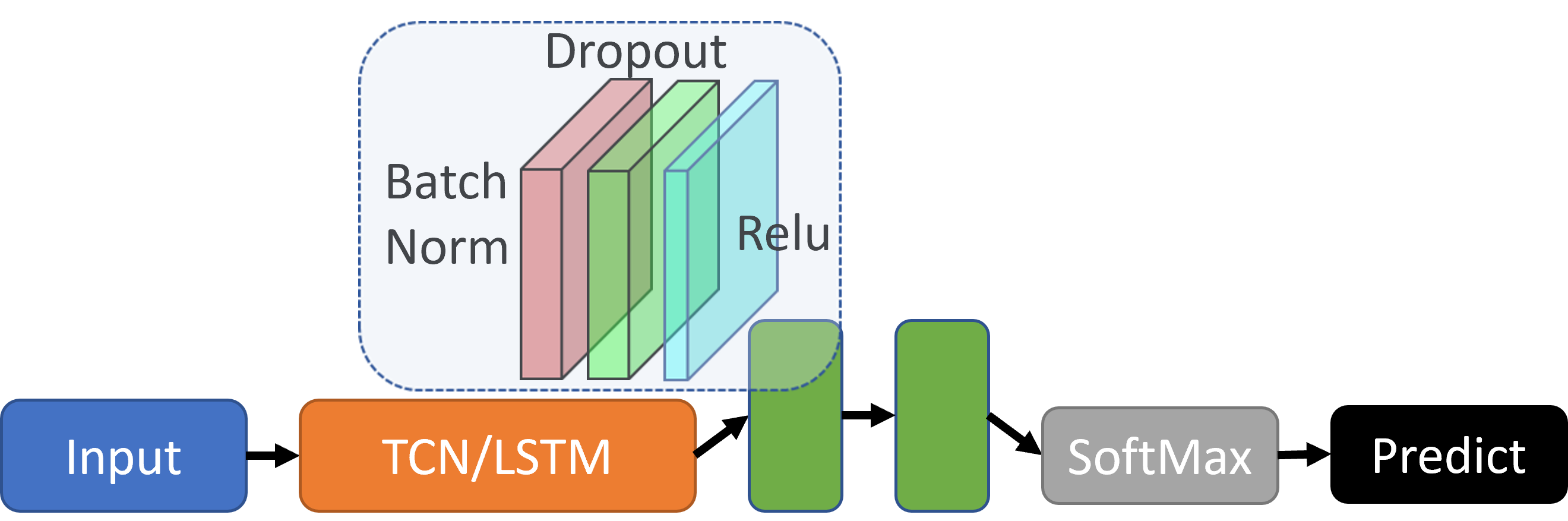}
			
		\end{center}
		\caption{The general architecture for our multi-class classification model.}
		\label{fig:single}
	\end{figure*}

	\section{Results}
	For our experiments, we split up the dataset into three folds. Using these folds, we perform 3-fold cross-validation to find accuracies, precision, and recall statistics. Note that only accuracies were provided in tables as other papers only include some metrics and not all of these metrics.


	\subsection{Analysis of multi-class binary classifiers}
	The binary classification model thrived with an LSTM over a TCN network. However, unlike a single classifier, the binary model performed better by having both polar and Cartesian coordinated for the body joints (method 5 of data normalization). Method 5 offers the highest performance, with a gesture recognition accuracy of 95.4\%.  The high accuracy reported in binary classifiers is due to the unique evaluation protocol.  Because of the imbalanced nature of binary classifiers in multi-class settings, accuracy reported in this form is greatly skewed towards accepting false positives.

	\begin{table}[!htb]
		\centering
		\caption{Multi-Class Binary Classifier Performance}\label{tab:onePerformance}
		\begin{tabular}{@{}ccc|ccc@{}}
			\multicolumn{3}{c|}{Properties}& \multicolumn{3}{c}{Accuracy (\%)} \\									
			Network	& Data Norm Method &Frames	&	Static	&	Dynamic	&	Average 	 \\
			\hline	
			LSTM & 3  & 256	 &94.6 	 & 	 93.1 	 & 	 93.8  \\
			LSTM & 3	 & 512 &	 94.2 	 & 	 93.9 	 & 94.0	 \\
			LSTM& 5 &	256  &95.4 & 	95.3	 & 	95.4   \\
			TCN & 1 & 32	&	93.2&	92.3&	93.1 	\\
			
		\end{tabular}
	\end{table}

	\subsection{Analysis of multi-class classifiers}
	Table \ref{tab:multiPerformance} shows the performance of our top four models for the multi-class classifier along with one of the TCN models to showcase how different deep learning methods can impact results. It can be seen that the LSTM is far more effective in classifying the data than the TCN model. As for the effect of the data pre-processing of the body joint information, our experiments show that the best results are achieved when using the angular data extraction method (method 3 and method 5). We experimented with frames 1,2,4,8,16...512 and found that a combination of 128 and 256 frames provides the highest performance.  Because the different actions and how each person performs each action varies widely, we created selected models that analyze shorter actions with 128 frames and longer actions with 256 frames and then combine their results together yielding a higher accuracy.
	
	We deduced the LSTM with Data Normalization Methods 3 and 5 (use of polar and rectangular coordinates), along with a combination of 128 and 256 frames, which seems to be the best multi-class classifier.  This combination yielded an accuracy of 70.8\% for gesture recognition on the PRAXIS dataset.
	
	\begin{table}[!htb]
		\centering
		\caption{Multi-Class Classifier Performance}\label{tab:multiPerformance}
		\begin{tabular}{@{}ccc|ccc@{}}
			\multicolumn{3}{c|}{Properties} 	&  \multicolumn{3}{c}{Accuracy (\%)} \\									
			Network	& Data Norm Method &Frames	&	Static	&	Dynamic	&	Average	 \\
			\hline	
			LSTM & 3/5 &128/256 	& 	74.3 	& 	67.3 	& 	70.8 \\
			LSTM & 3  	& 256	 	& 	65.3 	& 	57.2 	& 	61.3  \\
			LSTM & 3	 & 128 		&	67.0 	& 	61.2 	& 	64.1  \\
			LSTM& 5 &	256  		&	63.9 	& 	58.1	& 	61.0  \\
			TCN & 1 & 128			&	45.7	&	39.3	&	42.5 \\
			
		\end{tabular}
	\end{table}
	
	Table \ref{tab:performance} shows the results of our experiments against other approaches in \cite{Negin2018praxis}.  Our  approach achieved an accuracy of 70.8\% which outperforms other Skeleton joint-based approaches.
	
	\begin{table}
		\centering
		\caption{Hand Gesture Classification Performance}\label{tab:performance}
		\begin{tabular}{@{}c|ccc@{}}
			&  \multicolumn{3}{c}{Accuracy (\%)}  \\									
			Method		&	Static	&	Dynamic	&	Average  \\
			\hline	
			Proposed (Multi-class LSTM)	 & 	74.3 & 67.3 & 70.8	 \\
			
			\hline
			Skeleton (Distance) 				&	70.0 	& 	57.0	& 	63.5 \\
			Skeleton (Angle) 					&	57.2	&	51.4	&	54.3	\\	 
			Skeleton (Distance + Angle) 		&	61.8 	& 	55.8	& 	58.8  \\
			Multimodal Fusion (RGB + Skeleton)	&	72.4	&	62.8	&	67.6	\\
			Improved  Trajectories (MBHx/MBHy)	&	70.3	&	75.5	&	72.9	\\
			Deep Learning (CNN + LSTM)			&	92.9	&	76.6	&	84.7	\\ 
		\end{tabular}
	\end{table}
	
	\subsection{Analysis of multi-class binary classifiers}
	Similar to the multi-class classifier, the binary classification method also thrived with an LSTM over a TCN network. However, unlike the single classifier, the multi-classifier model performed better by having both polar and Cartesian coordinated for the skeletal joints (method 5 of data normalization). The best classifier among the binary classification options happens to be the LSTM network, with data normalization method 5 and 256 frames (row 3 of Table \ref{tab:multiPerformance}). This approach has the highest precision compared to all other models tested. We selected the best classifier by analyzing not just accuracy, since the data imbalances in the test sets for this classifier may have skewed the accuracies. Since the set is mostly comprised of actions being imperfect and rather abnormal, about 95\% of the data would come out with false labels. By using precision, we could accommodate for this, by counting the times when the machine classified the action to be true, and seeing how often it was correct in its guess.
	
	\begin{figure*}[!ht] 
		\begin{center} 
			\begin{tabular}{cc} 
				\includegraphics[width=0.45\textwidth]{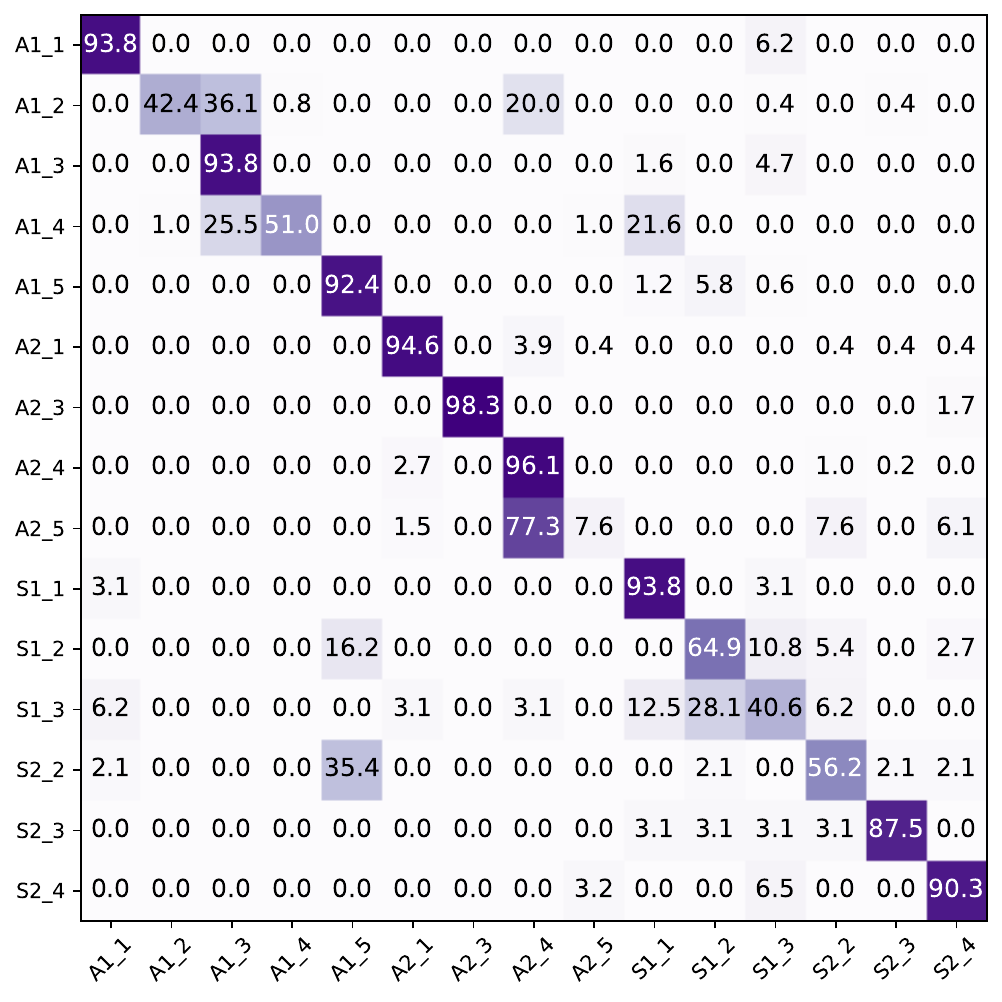}
				&
				\includegraphics[width=0.45\textwidth]{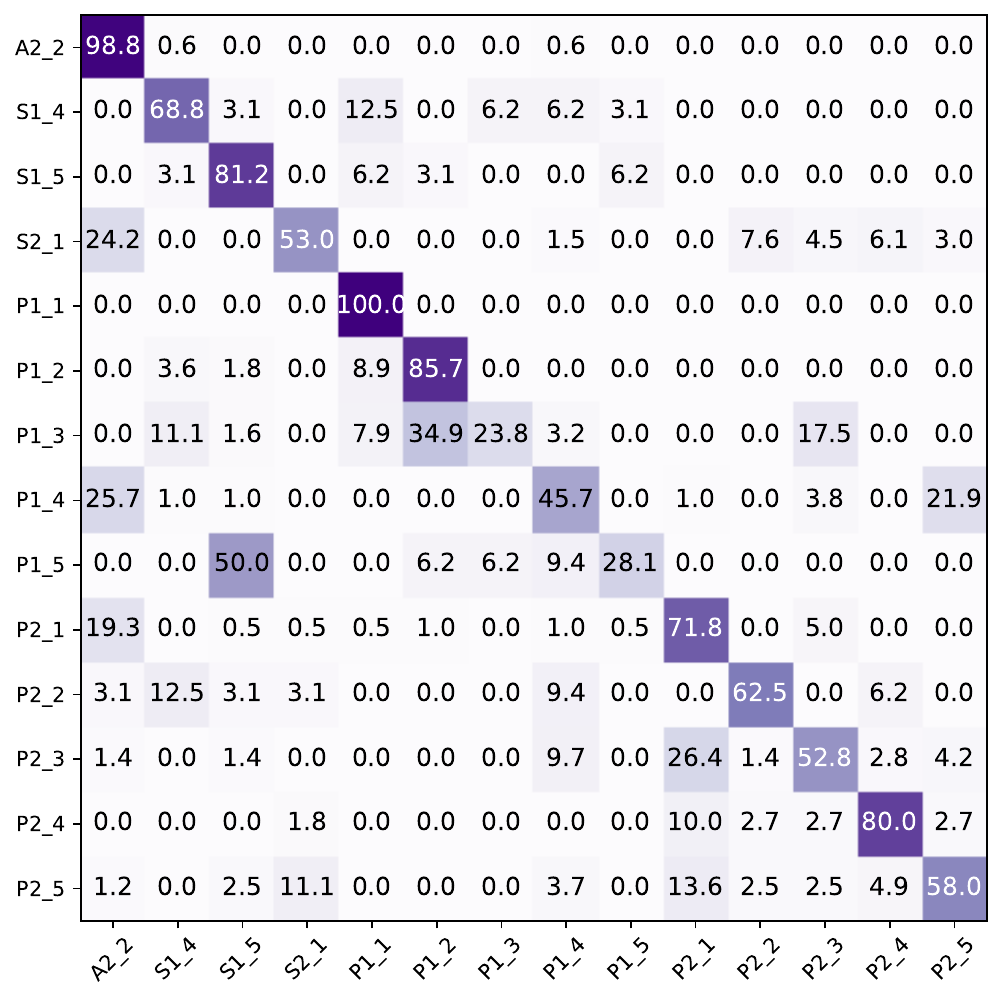} 
				\\
				$(a)$& $(b)$ \\
				
			\end{tabular}
		\end{center}
		\caption{Confusion matrix from the multi-class classifier using: 
			$(a)$ Static gestures; 
			$(b)$ Dynamic gestures.}			
		\label{fig:matrix}
	\end{figure*}
	
	\subsection{Static vs Dynamic gestures}
	
	An important observation based on the  numbers reported in  Table \ref{tab:performance} is that almost every classifier (including the baseline models) performed better on static gestures rather than on the dynamic ones. 
	
	Figure \ref{fig:matrix} shows the confusion matrices from the results of the multi-class LSTM. 7 out of the 15 gestures came out with individual accuracies of over 70\%, while only 3 out of 14 dynamic gestures have individual accuracies of over 70\%. This is likely because the temporal dependencies of the dynamic actions were much harder to understand for an LSTM compared to those for the static gestures.
	
	\subsection{Comparison of the Proposed and Existing Models}
	
	The performance  of the proposed classification approach reported in Table \ref{tab:performance} suggests that the  multi-class LSTM reached the best accuracy,  70.8\%.
	
	When comparing the multi-class classifier against the methods used by [14], our classifier is superior to all the other methods that rely on body joint data alone. It also achieved better results compared to the Multimodal Fusion method which uses both RGB and joint information in its classification model. When considering all methods that use all types of information, the deep learning method proposed in [14] achieves a superior accuracy of 84.7\% but requires RGB, depth, and joint information, thus greatly increasing the computational complexity.

	\section{Conclusion}
	
	This paper offers a machine learning approach that predicts hand gestures using only joint coordinate data. Using only joint data is greatly beneficial as it consumes less memory and requires much less computational power compared to data such as images. The data used for our experiment is the PRAXIS dataset which consists of different hand gestures performed by 60 individuals. Upon experimenting with different machine learning models to classify gestures using joint data, we developed a machine learning model using LSTM that achieves an accuracy of 70.8\% which is superior to other similar skeletal data based classifiers proposed in \cite{Negin2018praxis}. We notice an important pre-processing method that greatly assisted in improving the performance of the system is the use of polar coordinates to represent the data as opposed to Cartesian coordinates.  Another vital parameter for optimization is the duration of the sliding window.  In this paper, a long duration window such as 128 frames or 256 frames has been shown to yield the most optimal performance.  A major limitation of using only joint data is the limited amount of joints that can be used for learning.  Without joint orientation or a large number of joints, it is often difficult to tell the exact gesture such as open palm vs. fist which contains the same hand coordinates.

	\section{Future Works}
	
	Looking into the future, there are many ways we can improve our experiments. One improvement would be to try different deep learning architectures optimized for analyzing temporal space such as transformers.  Another aspect is the expansion of the dataset to include more subjects to reduce bias in the system as well as include additional joint points such that the orientation of the limbs can be learned.
	
		\section{Acknowledgement}
		{This research was partially supported by the Natural Sciences and Engineering Research Council of Canada (NSERC) through the grant ``Biometric-enabled Identity management and Risk Assessment for Smart Cities'' and by the Department of National Defense's Innovation for Defense Excellence and Security (IDEaS) program, Canada. R. Islam also acknowledges the Schulich School of Engineering Summer Studentship Award, University of Calgary.}

	{\small
		\bibliographystyle{IEEEtran}
		\bibliography{bias}
	}
	
\end{document}